\documentclass[11pt,a4paper]{article}
\usepackage[hyperref]{acl2019}
\usepackage{times}
\usepackage{latexsym}
\usepackage{dsfont}
\usepackage{url}
\usepackage{color}
\usepackage{booktabs} % For formal tables
\usepackage{CJKutf8}
\usepackage{graphicx}
\usepackage{amsmath}
\usepackage{multirow, makecell, caption}

\aclfinalcopy % Uncomment this line for the final submission
\def\aclpaperid{1183} %  Enter the acl Paper ID here

\title{Ensuring Readability and Data-fidelity using Head-modifier Templates \\ in Deep Type Description Generation}

\author{Jiangjie Chen\textsuperscript{\rm $\dagger$}, 
 Ao Wang\textsuperscript{\rm $\dagger$}, 
 Haiyun Jiang\textsuperscript{\rm $\dagger$}, 
 Suo Feng\textsuperscript{\rm $\dagger$}, 
 Chenguang Li\textsuperscript{\rm $\dagger$},
 Yanghua Xiao\textsuperscript{\rm $\dagger\ddagger$\thanks{\; Corresponding author: Yanghua Xiao.}} \\
\textsuperscript{\rm $\dagger$}Shanghai Key Laboratory of Data Science, School of Computer Science, Fudan University, China\\
\textsuperscript{\rm $\ddagger$}Shanghai Institute of Intelligent Electronics \& Systems, Shanghai, China\\
\{jiangjiechen14, awang15, jianghy16, fengs17, cgli17, shawyh\}@fudan.edu.cn\\
}

\date{}

\begin{document}
\maketitle

\begin{abstract}

A type description is a succinct noun compound which helps human and machines to quickly grasp the informative and distinctive information of an entity.
Entities in most knowledge graphs (KGs) still lack such descriptions, thus calling for automatic methods to supplement such information.
However, existing generative methods either overlook the grammatical structure or make factual mistakes in generated texts.
To solve these problems, we propose a head-modifier template-based method to ensure the readability and data fidelity of generated type descriptions.
We also propose a new dataset and two automatic metrics for this task.
Experiments show that our method improves substantially compared with baselines and achieves state-of-the-art performance on both datasets.

\end{abstract}

\section{Introduction}

%Humans understand the world by categorizing objects into concepts, which is the first step in human cognition from concrete to abstraction \cite{murphy2004big}.
%For human, the fastest way to grasp the essence of an unfamiliar object is to learn about its \textit{basic-level concept} \cite{murphy2004big}, which is an appropriate level of concepts.
%A basic-level concept is neither too general nor too specific, and must be \textit{informative} and \textit{distinctive} \cite{wang2015inference}.
%Such process is also called \textit{basic-level conceptualization} by psychologists \cite{murphy2004big}.
%For example, when asked about ``what is an \texttt{iPhone X}?'', most people first think of \texttt{Apple's smartphone} instead of \texttt{product} or \texttt{popular cellular wireless network phone}.

Large-scale open domain KGs such as DBpedia \cite{auer2007dbpedia}, Wikidata \cite{vrandevcic2014wikidata} and CN-DBpedia \cite{xu2017cn} are increasingly drawing the attention from both academia and industries, and have been successfully used in many applications that require background knowledge to understand texts.

In KGs, a \textbf{type description} \cite{Bhowmik2018EntityDescriptions} is a kind of description which reflects the rich information of an entity with little cognitive efforts.
A type description must be \textit{informative}, \textit{distinctive} and \textit{succinct} to help human quickly grasp the essence of an unfamiliar entity.
Compared to other kinds of data in a KG, \textit{types} in entity-typing task \cite{shimaoka2016neural,ren2016afet} are too general and not informative enough (e.g., when asked about ``what is \texttt{rue Cazotte}?'',  \texttt{street in Paris, France} is obviously more informative and distinctive than a type \textit{location}.), and the fixed type set is too inflexible to expand;
while \textit{infobox} and \textit{abstract} are too long with too much information, which increases cognitive burden.

Type descriptions are useful for a wide range of applications, including question answering (e.g. what is \textit{rue Cazotte}?),  named entity disambiguation (e.g. Apple (\textit{fruit of the apple tree}) vs Apple (\textit{American technology company})), taxonomy enrichment, etc.
However, many entities in current open-domain KGs still lack such descriptions.
For example, in DBpedia and CN-DBpedia respectively, there are only about 21\% and 1.8\% entities that are provided with such descriptions\footnote{According to DBpedia 2016-10 dump and CN-DBpedia 2015-07 dump.}.

\begin{figure}[t]
	\centering
	\includegraphics[width=\linewidth]{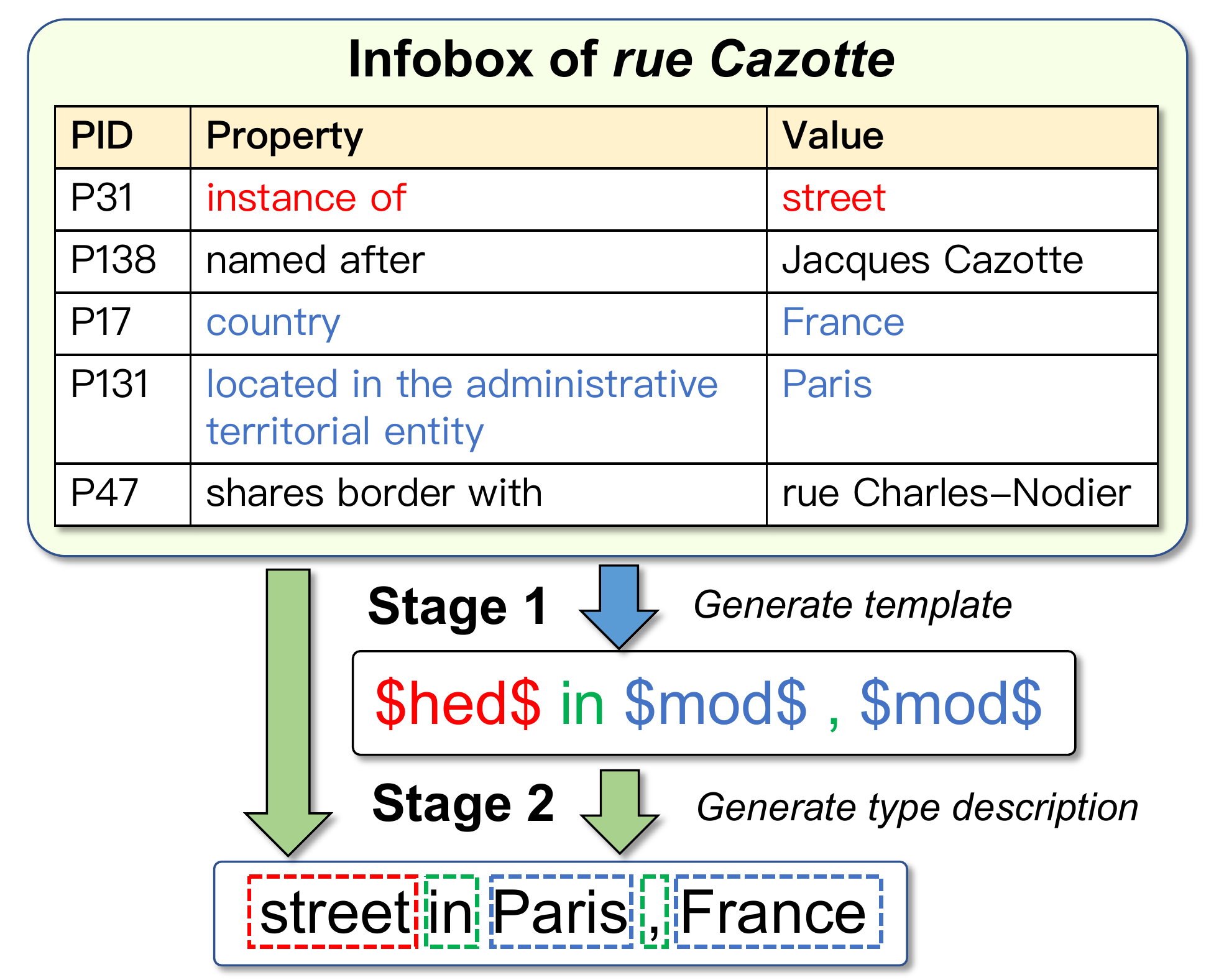}
	\caption{An example of the two-stage generation of our head-modifier template-based method. \texttt{\$hed\$} and \texttt{\$mod\$} are the placeholder for head and modifier components in the template.}
	\label{fig:front}
\end{figure}

Essentially, a type description is a noun compound, which follows a grammatical rule called \textit{head-modifier rule} \cite{hippisley2005head,wang2014head}. 
It always contains a \textit{head} component (also head words or heads), and usually contains a \textit{modifier} component (also modifier words or modifiers).
The head component representing the type information of the entity makes it \textit{distinctive} from entities of other types; the modifier component limits the scope of that type, making it more fine-grained and \textit{informative}. 
For example, in \texttt{street in Paris, France}, the head word \texttt{street} indicates that it is a \textit{street}, and the modifier words \texttt{Paris} and \texttt{France} indicate the street is located in \textit{Paris, France}.

Due to the low recall and limited patterns of extractive methods \cite{hearst1992automatic}, generative methods are more suitable to acquire more type descriptions.
Generally, there are several challenges in generating a type description from an infobox: 
1) it must be grammatically correct to be readable, given that a trivial mistake could lead to a syntax error (e.g. \texttt{street \textit{with} Paris, France}); 
2) it must guarantee the data fidelity towards input infobox, e.g., the system shouldn't generate \texttt{street in \textit{Germany}} for a French street; 
3) its heads must be the correct types for the entity, and a mistake in heads is more severe than in modifiers, e.g., in this case, \texttt{\textit{river} in France} is much worse than \texttt{street in \textit{Germany}}.

We argue that the head-modifier rule is crucial to ensure readability and data-fidelity in type description generation. However, existing methods pay little attention to it. 
\citet{Bhowmik2018EntityDescriptions} first propose a dynamic memory-based generative network to generate type descriptions from infobox in a neural manner. 
They utilize a memory component to help the model better remember the training data.
However, it tends to lose the grammatical structure of the output, as it cannot distinguish heads from modifiers in the generation process.
Also, it cannot handle the out-of-vocabulary (OOV) problem, and many modifier words may be rare and OOV.
Other data-to-text \cite{Wiseman:2017dp,sha2018order} and text-to-text \cite{gu2016incorporating,gulcehre2016pointing,See:2017wf} models equipped with copy mechanism alleviate OOV problem, without considering the difference between heads and modifiers, resulting in grammatical or factual mistakes.

To solve the problems above, we propose a head-modifier template-based method.
To the best of our knowledge, we are the first to integrate head-modifier rule into neural generative models.
Our method is based on the observation that a head-modifier template exists in many type descriptions.
For example, by replacing heads and modifiers with placeholders \texttt{\$hed\$} and \texttt{\$mod\$}, the template for \texttt{street in Paris, France} is \texttt{\$hed\$ in \$mod\$, \$mod\$}, which is also the template for a series of similar type descriptions such as \texttt{library in California, America}, \texttt{lake in Siberia, Russia}, etc.
Note that, the \texttt{\$hed\$} and \texttt{\$mod\$} can appear multiple times, and punctuation like a comma is also an important component of a template.

Identifying the head and modifier components is helpful for providing structural and contextual cues in content selection and surface realization in generation, which correspond to data fidelity and readability respectively. 
As shown in Fig.\ref{fig:front}, the model can easily select the corresponding properties and values and organize them by the guidance of the template.
%For example, given property-value pairs \texttt{(occupation, singer)} and \texttt{(country, America)}, the model learns that the former is more likely to appear in \texttt{\$hed\$}.
The head-modifier template is universal as the head-modifier rule exists in any noun compound in English, even in Chinese \cite{hippisley2005head}.
Therefore, the templates are applicable for open domain KGs, with no need to design new templates for entities from other KGs.

There are no existing head-modifier templates to train from, therefore we use the dependency parsing technique \cite{manning2014stanford} to acquire templates in training data. 
Then, as presented in Fig.\ref{fig:front}, our method consists of two stages: 
in Stage 1, we use an encoder-decoder framework with an attention mechanism to generate a template;
in Stage 2, we use a new encoder-decoder framework to generate a type description, and reuse previously encoded infobox and apply a copy mechanism to preserve information from source to target.
Meanwhile, we apply another attention mechanism upon generated templates to control the output's structure.
We then apply a context gate mechanism to dynamically select contexts during decoding.

In brief, our contributions\footnote{https://github.com/Michael0134/HedModTmplGen} in this paper include,
1) we propose a new head-modifier template-based method to improve the readability and data fidelity of generating type descriptions, which is also the first attempt of integrating head-modifier rule into neural generative models;
2) we apply copy and context gate mechanism to enhance the model's ability of choosing contents with the guidance of templates;
3) we propose a new dataset with two new automatic metrics for this task, and experiments show that our method achieves state-of-the-art performance on both datasets.

%In brief, our contributions\footnote{We will release the code and dataset after paper's acceptance.} are listed as follows:
%\begin{itemize}
%	\item We propose a new head-modifier template-based method to improve the readability and data fidelity of type descriptions generation, which is also the first attempt of integrating head-modifier rule into deep generative models. 
%	\item We apply a copy mechanism and context gate mechanism to enhance the model's ability of choosing contents with the guidance of the templates.
%	\item We also propose a new dataset with two new metrics for this task, and experiments show that our method achieves state-of-the-art performance over competitive baselines on both datasets.
%\end{itemize}

\begin{figure*}[!hbt]
	\centering
	\includegraphics[width=0.9\linewidth]{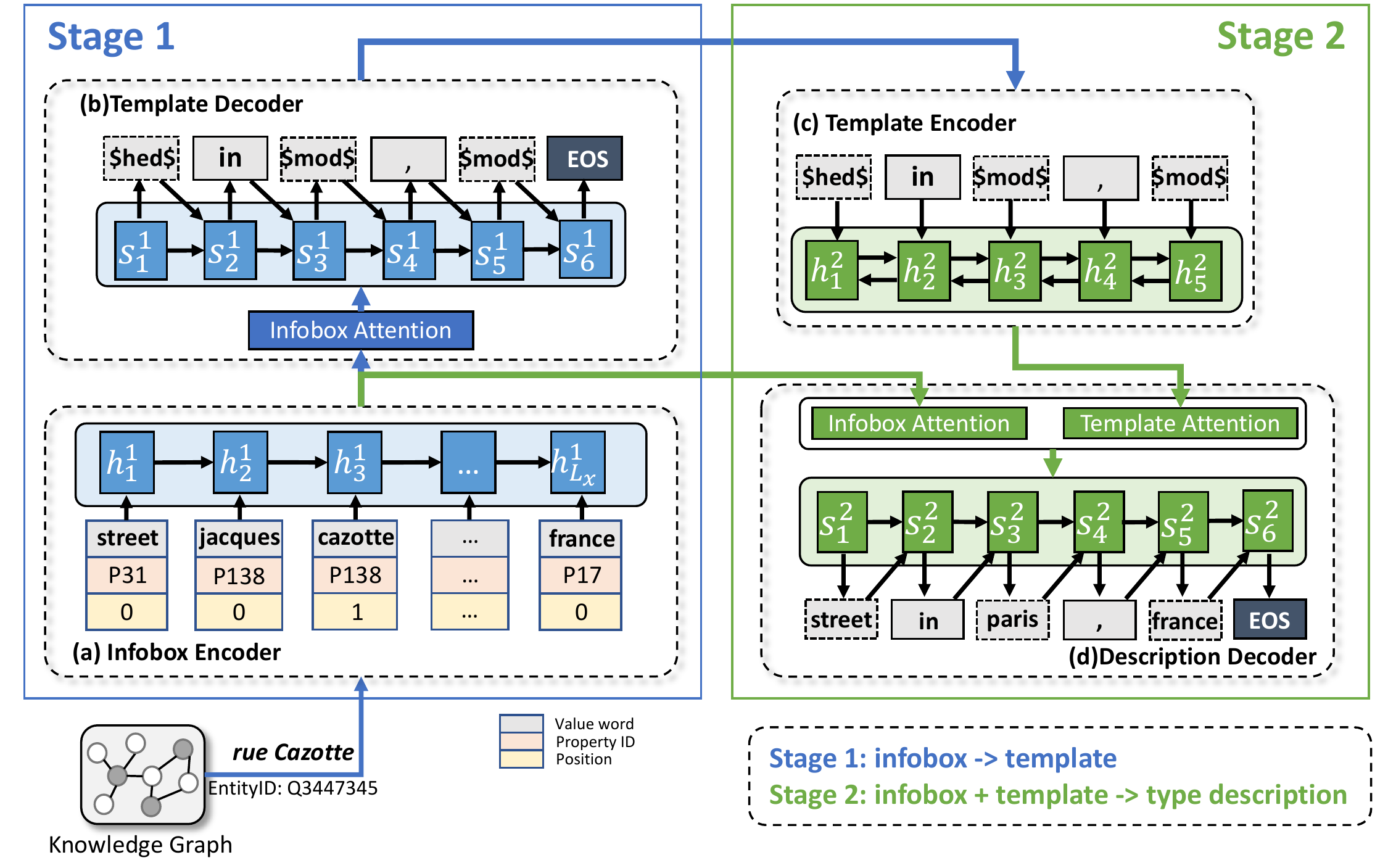}
	\caption{Overall architecture of our method. In Stage 1, the model generates a template from infobox of entity \texttt{rue Cazotte} (the entity can be found at Wikidata by EntityID), then in Stage 2 the model completes this template by reusing the infobox and generates a type description for this entity. }
	\label{fig:framework}
\end{figure*}

\section{Method}

In this section, we demonstrate our method in detail. As shown in Fig.\ref{fig:framework}, given an entity from Wikidata\footnote{www.wikidata.org} and its corresponding infobox, we split the generation process into two stages. 
In Stage 1, the model takes as input an infobox and generates a head-modifier template.
In Stage 2, the model takes as input the previously encoded infobox and the output template, and produces a type description.
Note that our model is trained in an end-to-end manner.

\subsection{Stage 1: Template Generation}

In this stage, we use an encoder-decoder framework to generate a head-modifier template of the type description.

\subsubsection{Infobox Encoder} \label{encoder}

\begin{figure}[htp]
	\centering
	\includegraphics[width=\linewidth]{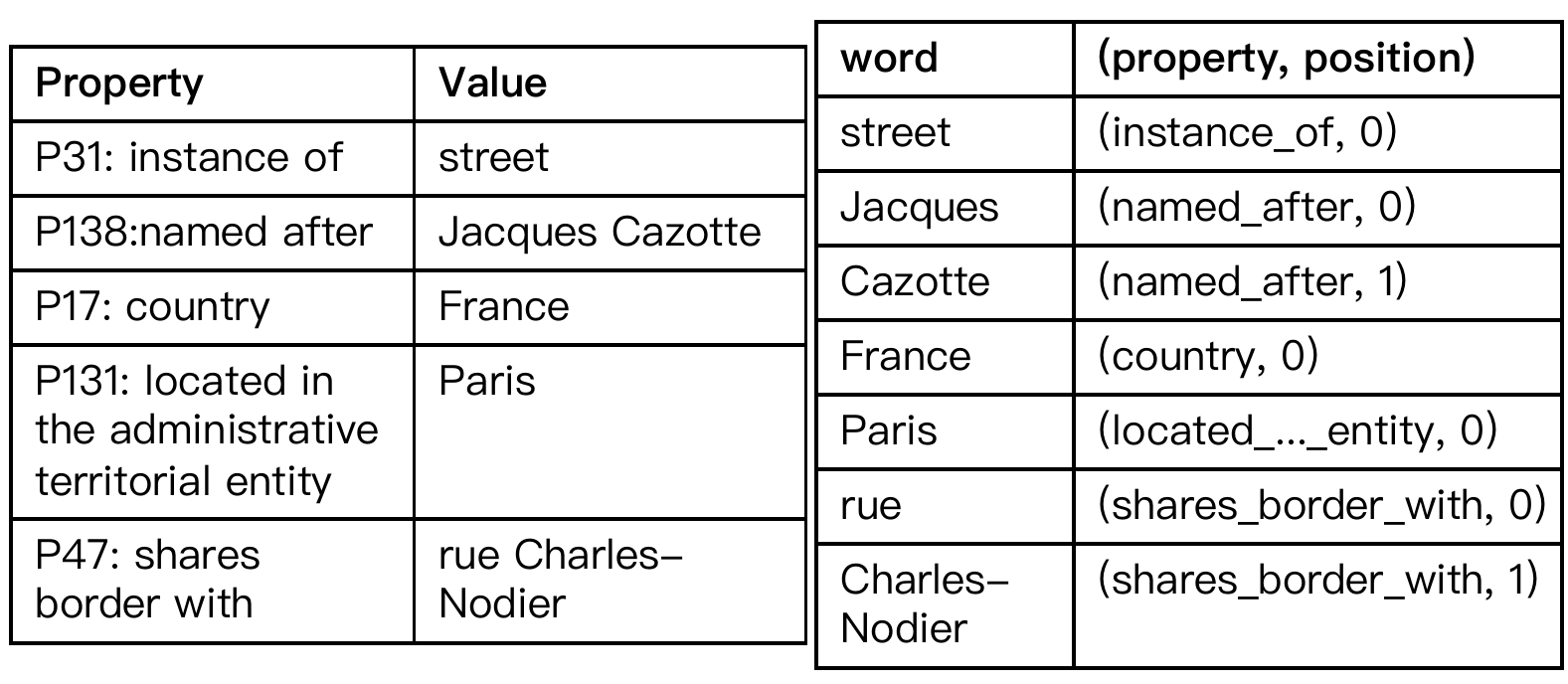}
	\caption{An example of reconstructing a Wikidata infobox (left) into a sequence of words with property and position information (right). \texttt{PN} denotes a property ID in Wikidata.}
	\label{fig:infobox}
\end{figure}

Our model takes as input an infobox of an entity, which is a series of (\texttt{property}, \texttt{value}) pairs denoted as $\mathcal{I}$. We then reconstruct them into a sequence of words to apply Seq2Seq learning. 
In order to embed structural information from the infobox into word embedding $\textbf{x}_i$, following \citet{lebret2016neural}, we represent $\textbf{x}_i = [\textbf{v}_{x_i};\textbf{f}_{x_i};\textbf{p}_{x_i}]$ for the \texttt{i}-th word $x_i$ in the values, with the word embedding $\textbf{v}_{x_i}$ for ${x_i}$, a corresponding property embedding $\textbf{f}_{x_i}$ and the positional information embedding $\textbf{p}_{x_i}$, and $[\cdot ;\cdot ]$ stands for vector concatenation.

For example, as shown in Fig.\ref{fig:infobox}, we reconstruct (\texttt{named after}, \texttt{Jacques Cazotte}) into \texttt{Jacques} with (\texttt{named\_after}, \texttt{0}) and \texttt{Cazotte} with (\texttt{named\_after}, \texttt{1}), as \texttt{Jacques} is the first token in the value and \texttt{Cazotte} is the second.
Next, we concatenate the embedding of \texttt{Jacques}, \texttt{named\_after} and \texttt{0} as the reconstructed embedding for \texttt{Jacques}. 
Notice that, we have three separate embedding matrices for properties, value words and position, that is, even though the property \texttt{country} is the same string as the value \texttt{country}, they are not the same token. 

Then, we employ a standard GRU \cite{chung2014empirical} to read the input $\textbf{X}=\{\textbf{x}_i\}_{i=1}^{L_x}$, then produce a sequence of hidden states $\textbf{H}_x = \{\textbf{h}_i^1\}_{i=1}^{L_x}$, which are shared in both stages, where $L_x$ is the length of the input sequence.

\subsubsection{Template Annotation}

\begin{figure}[htp]
	\centering
	\includegraphics[width=\linewidth]{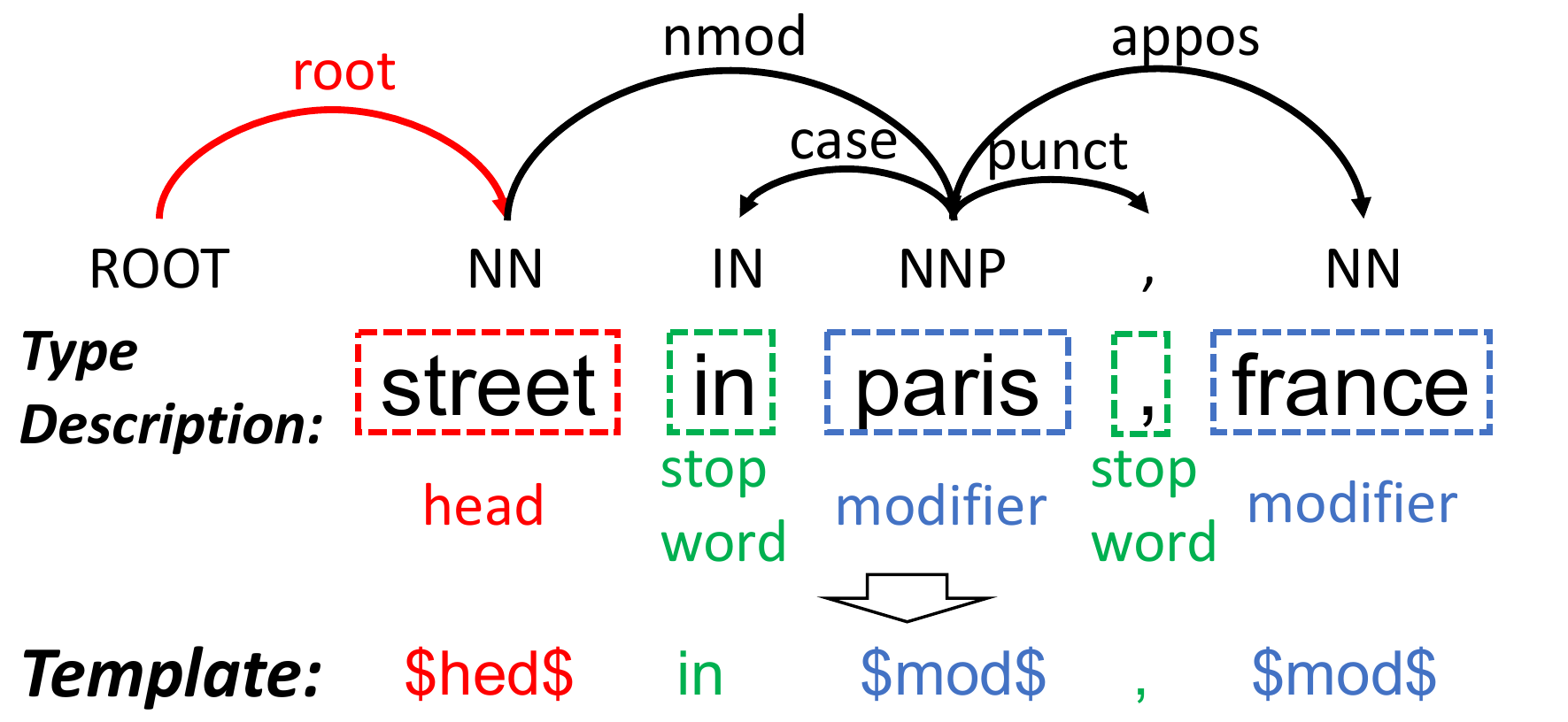}
	\caption{An example of extracting head-modifier template from type description by dependency parsing using Stanford CoreNLP toolkit.}
	\label{fig:tmpl}
\end{figure}

In this task, the type descriptions are diversified yet following the head-modifier rule. 
The Stage 1 in our model learns the templates from training data, but there are no existing templates for the template generation training.
Therefore, we acquire head-modifier templates by using a dependency parser provided by Stanford CoreNLP \cite{manning2014stanford}.

Specifically, a type description is formed by head words (or heads), modifier words (or modifiers) and conjunctions.
In our work, we refer to words that are types as heads in a type description, so there could be multiple heads.
For example, \texttt{singer} and \texttt{producer} in \texttt{American singer, producer} are both head words.

During dependency parsing, the \texttt{root} of a noun compound is always a head word of the type description. 
Therefore, we acquire heads by finding the root and its parallel terms.
The remaining words except conjunctions and stopwords are considered to be modifiers. 
We then obtain the template by substituting heads with \texttt{\$hed\$} and modifiers with \texttt{\$mod\$}, as shown in Fig.\ref{fig:tmpl}.

\subsubsection{Template Decoder} \label{dec1}

In template generation, the template decoder $\mathcal{D}_1$ takes as input the previous encoded hidden states $\textbf{H}_x$ and produces a series of hidden states $\{\textbf{s}_1^1, \textbf{s}_2^1, ..., \textbf{s}_{L_x}^1\}$ and a template sequence $\mathcal{T} = \{t_1, t_2, ..., t_{L_t}\}$, where $L_t$ is the length of the generated template.
As template generation is a relatively lighter and easier task, we apply a canonical attention decoder as $\mathcal{D}_1$, with GRU as the RNN unit. 

Formally, at each time step $j$, the decoder produces a context vector $\textbf{c}_j^1$,
\begin{equation}
	\textbf{c}_j^1 = \sum_{i=1}^{L_x} \alpha_{ij}\textbf{h}_i^1;
	\alpha_{ij} = \frac{\eta (\textbf{s}_{j-1}^1,{\textbf{h}^1_i})}{\sum_{k=1}^{L_x} \eta (\textbf{s}_{j-1}^1,{\textbf{h}^1_i}))}
\end{equation}
where $\eta (\textbf{s}_j^1, \textbf{h}_i^1)$ is a relevant score between encoder hidden state $\textbf{h}_i^1$ and a decoder hidden state $\textbf{s}_j^1$. 
Among many ways to compute the score, in this work, we apply general product \cite{luong2015effective} to measure the similarity between both:
\begin{equation}
	\eta (\textbf{h}_i^1, \textbf{s}_{j-1}^1) = \textbf{h}_i^{1\top} \textbf{W}_1 \textbf{s}_{j-1}^1
\end{equation} 
where $\textbf{W}_1$ is a learnable parameter.

Then the decoder state is updated by $\textbf{s}_j^1=GRU([t_{j-1};\textbf{c}_j^1],\textbf{s}_{j-1}^1)$. 
Finally, the results are fed into a softmax layer, from which the system produces $t_j$.

\subsection{Stage 2: Description Generation}

After Stage 1 is finished, the generated template sequence $\mathcal{T}$ and the infobox encoder hidden states $\textbf{H}_x$ are fed into Stage 2 to produce the final type description.

\subsubsection{Template Encoder}

As the template is an ordered sequence, we use a bidirectional \cite{schuster1997bidirectional} GRU to encode template sequence into another series of hidden states $\textbf{H}_t = \{\textbf{h}_i^2\}_{i=1}^{L_t}$. 
Then we fed both $\textbf{H}_t$ and $\textbf{H}_x$ to the description decoder for further refinement.

\subsubsection{Description Decoder}

The description decoder $\mathcal{D}_2$ is a GRU-based decoder, which utilizes a dual attention mechanism: a canonical attention mechanism and a copy mechanism to attend over template representation $\textbf{H}_t$ and infobox representation $\textbf{H}_x$ respectively. 
This is because we need the model to preserve information from the source while maintaining the head-modifier structure learned from the templates. 

In detail, let $\textbf{s}_j^2$ be $\mathcal{D}_2$'s hidden state at time step $j$. The first canonical attention mechanism is similar to the one described in Section \ref{dec1}, except that the decoder hidden states are replaced and related learnable parameters are changed.
By applying this, we obtain a context vector $\textbf{c}^t_j$ of $\textbf{H}_t$ and a context vector $\textbf{c}^x_j$ of $\textbf{H}_x$.

Then, we use context gates proposed by \citet{tu2017context} to dynamically balance the contexts from infobox, template, and target, and decide the ratio at which three contexts contribute to the generation of target words. 

Formally, we calculate the context gates $g_j^*$ by
\begin{equation}
	\begin{split}
		g_j^x &= \sigma (\textbf{W}_g^xe(y_{j-1}) + \textbf{U}_g^x\textbf{s}_{j-1} + \textbf{C}_g^x\textbf{c}_j^x)\\
		g_j^t &= \sigma (\textbf{W}_g^te(y_{j-1}) + \textbf{U}_g^t\textbf{s}_{j-1} + \textbf{C}_g^t\textbf{c}_j^t)
	\end{split}
\end{equation}
where $\textbf{W}_g^*, \textbf{U}_g^*, \textbf{C}_g^*$ are all learnable parameters, $\sigma$ is a sigmoid layer, and $e(y)$ embeds the word $y$.
After that, we apply a linear interpolation to integrate these contexts and update the decoder state:
\begin{equation}
\begin{split}
	\textbf{c}_j^2 = &(1-g_j^x-g_j^t)(\textbf{W}e(y_{j-1}) + \textbf{U}\textbf{s}_{j-1}^2)+ \\
	&g_j^x\textbf{C}_1\textbf{c}_j^x + g_j^t\textbf{C}_2\textbf{c}_j^t\\
	\textbf{s}_j^2 = &GRU([y_{j-1}; \textbf{c}_j^2], \textbf{s}_{j-1}^2)
\end{split}
\end{equation}
where $\textbf{W}, \textbf{U}, \textbf{C}_1, \textbf{C}_2$ are all learnable parameters.

To conduct a sort of slot filling procedure and enhance the model's ability of directly copying words from infobox, we further apply conditional copy mechanism \cite{gulcehre2016pointing} upon $\textbf{H}_x$.
As the produced words may come from the vocabulary or directly from the infobox , we assume a new decoding vocabulary $\mathcal{V'} = \mathcal{V} \cup \{x_i\}_{i=1}^{L_x} $, where $\mathcal{V}$ is the original vocabulary with the vocabulary size of $N$, and \texttt{unk} is the replacement for out-of-vocabulary words.

Following \citet{Wiseman:2017dp}, the probabilistic function of $y_j$ is as follows:
\begin{equation}
	\begin{split}
	&p(y_j, z_j|y_{<j}, \mathcal{I}, \mathcal{T})= \\
		&\begin{cases}
			p_{copy}(y_j|y_{<j}, \mathcal{I}, \mathcal{T}) p(z_j|y_{<j},\mathcal{I}), &z_j=0 \\
			p_{gen}(y_j|y_{<j},\mathcal{I},\mathcal{T}) p(z_j|y_{<j},\mathcal{I}), &z_j=1
		\end{cases}
	\end{split}
\end{equation}
where $z_j$ is a binary variable deciding whether $y_j$ is copied from $\mathcal{I}$ or generated, and $p(z_j|\cdot)$ is the switcher between copy and generate mode which is implemented as a multi-layer perceptron (MLP). 
$p_{copy}(y_j|\cdot)$ and $p_{gen}(y_j|\cdot)$ are the probabilities of copy mode and generate mode respectively, which are calculated by applying softmax on copy scores $\phi_{copy}$ and generation scores $\phi_{gen}$. 
These scores are defined as follows:
\begin{equation}
	\begin{split}
		\phi_{gen} (y_j=v) &= \textbf{W}_g[\textbf{s}_j^2;\textbf{c}_j^2], v \in \mathcal{V}\cup\{\texttt{unk}\}\\
		\phi_{copy} (y_j=x_i) &= \tanh (\textbf{h}_i^x\textbf{W}_c)\textbf{s}_j^2, x_i \in \mathcal{V'}-\mathcal{V}
	\end{split}
\end{equation}
where $W_c, W_g$ are both learnable parameters. 
Therefore, a word is considered as a copied word if it appears in the value portion of the source infobox.

\subsection{Learning}

Our model is able to be optimized in an end-to-end manner and is trained to minimize the negative log-likelihood of the annotated templates $\mathcal{T}$ given infobox $\mathcal{I}$ and the ground truth type descriptions given $\mathcal{T}$ and $\mathcal{I}$. 
Formally,
\begin{equation}
\begin{split}
	\mathcal{L}_1 &= - \sum_{i=1}^{L_t} \log p(t_i|t_{<i},\mathcal{I})  \\
	\mathcal{L}_2 &= - \sum_{i=1}^{L_y} \log p(y_i|y_{<i},\mathcal{I}, \mathcal{T}) \\
	\mathcal{L} &= \mathcal{L}_1 + \mathcal{L}_2
\end{split}
\end{equation}
where $\mathcal{L}_1$ is the loss in Stage 1, $\mathcal{L}_2$ is the loss in Stage 2, and $L_y$ is the length of the target.

\section{Experiments}

In this section, we conduct several experiments to demonstrate the effectiveness of our method. 

\subsection{Datasets}

We conduct experiments on two English datasets sampled from Wikidata, which are referred to as \textbf{Wiki10K} and \textbf{Wiki200K} respectively. \textbf{Wiki10K} is the original dataset proposed by \citet{Bhowmik2018EntityDescriptions}, which is sampled from Wikidata and consists of 10K entities sampled from the official RDF exports of Wikidata dated 2016-08-01. 
However, this dataset is not only too small to reveal the subtlety of models, 
but it's also relatively imbalanced with too many \texttt{human} entities based on the property \texttt{instance of}.
Therefore, we propose a new and larger dataset \textbf{Wiki200K}, which consists of 200K entities more evenly sampled from Wikidata dated 2018-10-01.
Note that, in both \textbf{Wiki10K} and \textbf{Wiki200K}, we filter all the properties whose data type are not \texttt{wikibase-item}, \texttt{wikibase-property} or \texttt{time} according to Wikidata database reports\footnote{https://www.wikidata.org/wiki/Wikidata:Database\_reports/\\List\_of\_properties/all}.

KGs such as Wikidata are typically composed of semantic triples. 
A semantic triple is formed by a subject, a predicate, and an object, corresponding to entity, property and value in Wikidata.
We make sure that every entity from both datasets has at least 5 property-value pairs (or \textit{statement} in Wikidata parlance) and an English type description. 
The basic statistics of the two datasets are demonstrated in Table \ref{table:dstable}. 
Then, we randomly divide two datasets into train, validation and test sets by the ratio of 8:1:1.

\begin{table}[!h]
    \centering
	\begin{tabular}{ccc}
		\toprule
        \textbf{Datasets}&\textbf{Wiki10K}&\textbf{Wiki200K}\\
        \midrule
        \# entities&10,000 & 200,000\\
        \# properties & 480 & 900 \\
        vocabulary size&28,785&130,686\\
        \# \textit{avg} statement &8.90&7.96\\
%        Absolute Copy(\%) & 58.66 & 57.33 \\
        Copy(\%) & 88.24 & 71.30\\
        \bottomrule
	\end{tabular}
	\caption{Statistics for both datasets, where ``\#'' denotes the number counted, and \textit{avg} is short for \textit{average}. ``Copy(\%)'' denotes the copy ratio in the golden type descriptions excluding stopwords, which is similar to the metric \textbf{ModCopy} defined in Section \ref{metric}.}
	\label{table:dstable}
\end{table}

\begin{table*}[!hbtp]
	\centering
	\begin{tabular}{cccccccc}
		\toprule
		\multicolumn{8}{c}{\textbf{Wiki10K}}\\
		\hline
		\textbf{Model} & \textbf{B-1} & \textbf{B-2} & \textbf{RG-L} & \textbf{METEOR} & \textbf{CIDEr} & \textbf{ModCopy} & \textbf{HedAcc}  \\
		\hline
		AttnS2S & 53.96 & 47.56 & 55.25 & 29.95 & 2.753 & 69.45 & 52.82  \\
		% step: 10000
		Ptr-Gen& 64.24  & 57.11 & 65.37 & 36.42 & 3.536 & 83.88 &67.92 \\
		Transformer& 61.63 & 54.93 & 63.14 & 35.01  & 3.400 & 75.37 &61.13  \\
		DGN & 63.24 & 57.52 & 64.50 & 35.92 & 3.372 & 77.53 & 64.65  \\
		\textbf{Our work} & \textbf{65.09} & \textbf{58.72} & \textbf{66.92} & \textbf{37.55} & \textbf{3.717} &\textbf{86.04} &\textbf{70.68}  \\
		\midrule
		\multicolumn{8}{c}{\textbf{Wiki200K}}\\
		\hline
		\textbf{Model} & \textbf{B-1} & \textbf{B-2} & \textbf{RG-L} & \textbf{METEOR} & \textbf{CIDEr} & \textbf{ModCopy}& \textbf{HedAcc} \\
		\hline
		AttnS2S & 66.15 & 61.61 & 70.55 & 37.65 & 4.105 & 49.59 & 79.76 \\
		% step 
		Ptr-Gen& 70.13 & 66.21 & 75.21 & 41.38 & 4.664 &\textbf{58.27} &85.38  \\
		% step 
		Transformer& 69.78 & 66.07 & 75.60 & 41.52 & 4.654 & 53.85& 85.55  \\
		% step 
		DGN & 62.60 & 57.86  & 69.30& 34.84 & 3.815 & 48.30 & 81.31  \\
		% ep
		\textbf{Our work} & \textbf{73.69} & \textbf{69.59} & \textbf{76.77} & \textbf{43.54} & \textbf{4.847} & 58.14 &\textbf{85.81}  \\
		\bottomrule
	\end{tabular}
	\caption{Evaluation results of different models on both datasets.}
	\label{table:evaltable}
\end{table*}

\subsection{Evaluation Metrics} \label{metric}

Following the common practice, we evaluate different aspects of the generation quality with automatic metrics broadly applied in many natural language generation tasks, including BLEU (B-1, B-2) \cite{papineni2002bleu}, ROUGE (RG-L) \cite{lin2004rouge}, METEOR \cite{banerjee2005meteor} and CIDEr \cite{vedantam2015cider}. 
BLEU measures the n-gram overlap between results and ground truth, giving a broad point of view regarding fluency, while ROUGE emphasizes on the precision and recall between both.
METEOR matches human perception better and CIDEr captures human consensus.

Nonetheless, these metrics depend highly on the comparison with ground truth, instead of the system's input.
In this task, the output may still be correct judging by input infobox even if it's different from the ground truth.
Therefore, we introduce two simple automatic metrics designed for this task to give a better perspective of the data fidelity of generated texts from the following aspects:

\begin{itemize}
	\item \textbf{Modifier Copy Ratio (ModCopy)}. 
		We evaluate the data fidelity regarding preserving source facts by computing the ratio of modifier words (that is, excluding stopwords and head words) in the type descriptions that are copied from the source.
		In detail, we roughly consider a word in a type description as a copied word if it shares a \textit{L}-character (4 in our experiments) prefix with any word but stopwords in the values of source infobox. 
		For example, modifier \texttt{Japanese} could be a copied modifier word from the fact (\texttt{country}, \texttt{Japan}). 
		To clarify, the copy ratio of a type description can be calculated by $\frac{\# copied\_words}{\# all\_words - \# stopwords}$.
		%Formally, the fuzzy copy ratio are calculated by 
		%% TODO
		%\begin{equation}
		%	CopyRatio=\frac{\sum_{w \in C - S} \#w }{\sum_{w \notin S}\# w}, w \in TypeDesc
		%\end{equation}
		%where $C$ denotes $S$ denotes a set of stopwords. 
		The \textit{Modifier Copy Ratio} measures to what extent the informative words are preserved in the modifiers of the model's output.

	\item \textbf{Head Accuracy (HedAcc)}. 
		For a type description, it is crucial to make sure that the head word is the right type of entity. 
		Therefore, in order to give an approximate estimate of the data fidelity regarding head words, we also evaluate the head word's accuracy in the output.
		Note that aside from ground truth, infobox is also a reliable source to provide candidate types.
		Specifically, in Wikidata, the values in \texttt{instance of} (P31) and \texttt{subclass of} (P279) are usually suitable types for an entity, though not every entity has these properties and these types could be too coarse-grained like \texttt{human}.
		Therefore, after dependency parsing, we count the head words in the output with heads from corresponding ground truth and values of corresponding infobox properties, then gives an accuracy of the heads of output.
		%Formally, accuracy is calculated by 
		%%TODO
		%\begin{equation}
		%	HedAcc = \frac{\# h \in \{hs\}_{gold}\cup{v}_{P31\cup P279}}{\# head}
		%\end{equation}
		The \textit{Head Accuracy} measures model's ability of predicting the right type of the entity.
\end{itemize}

\subsection{Baselines and Experimental Setup} \label{baseline}

We compared our method with several competitive generative models.
All models except DGN are implemented with the help of OpenNMT-py \cite{opennmt}.
Note that we use the same infobox reconstructing method described in Section \ref{encoder} to apply Seq2Seq learning for all models except DGN since it has its own encoding method.
The baselines include:

\begin{itemize}
	\item \textbf{AttnSeq2Seq} \cite{luong2015effective}. AttnS2S is a standard RNN-based Seq2Seq model with an attention mechanism.
	\item \textbf{Pointer-Generator} \cite{See:2017wf}. Ptr-Gen is originally designed for text summarization, providing a strong baseline with a copy mechanism.
	Note that, in order to make a fairer comparison with our model, we additionally equip Ptr-Gen with context gate mechanism so that it becomes a no-template version of our method.
	\item \textbf{Transformer} \cite{vaswani2017attention}. Transformer recently outperforms traditional RNN architecture in many NLP tasks, which makes it also a competitive baseline, even if it's not specifically designed for this task.
	\item \textbf{DGN} \cite{Bhowmik2018EntityDescriptions}. DGN uses a dynamic memory based network with a positional encoder and an RNN decoder. It achieved state-of-the-art performance in this task.
\end{itemize}

% \textbf{AttnSeq2Seq} \cite{luong2015effective}. AttnS2S is a standard RNN-based Seq2Seq model with an attention mechanism.

% \textbf{Pointer-Generator} \cite{See:2017wf}. Ptr-Gen is originally designed for text summarization, providing a strong baseline with a copy mechanism.
% Note that, in order to make a fairer comparison with our model, we additionally equip Ptr-Gen with context gate mechanism so that it becomes a no-template version of our method.

% \textbf{Transformer} \cite{vaswani2017attention}. Transformer recently outperforms traditional RNN architecture in many NLP tasks, which makes it also a competitive baseline, even if it's not specifically designed for this task.

% \textbf{DGN} \cite{Bhowmik2018EntityDescriptions}. DGN uses a dynamic memory based network with a positional encoder and an RNN decoder. It achieved state-of-the-art performance in this task.

In experiments, we decapitalize all words and keep vocabularies at the size of 10,000 and 50,000 for \textbf{Wiki10K} and \textbf{Wiki200K} respectively, and use \texttt{unk} to represent other out-of-vocabulary words.

For the sake of fairness, the hidden size of RNN (GRU in our experiments) and Transformer in all models are set to 256. 
The word embedding size is set to 256, and the property and position embedding sizes are both set to 128. 
During training, we use Adam \cite{kingma2014adam} as the optimization algorithm.

\subsection{Results and Analysis}

The experimental results of metrics described in Section \ref{metric} are listed in Table \ref{table:evaltable}. 
In general, our method achieves state-of-the-art performance over proposed baselines.

As shown in the table, our method improves substantially compared with standard encoder-decoder models (AttnS2S and Transformer) and the previous state-of-the-art method (DGN).
Interestingly, DGN is out-performed by Ptr-Gen in \textbf{Wiki10K} and by most of the models in the larger dataset \textbf{Wiki200K}. 
We also notice that Transformer performs much better on \textbf{Wiki200K}, which is most likely because of its learning ability through massive training data. 
These results further prove the necessity of proposing our new dataset. 
Among baselines, Ptr-Gen achieves relatively better results due to copy mechanism and context gate mechanism.
These mechanisms give the model the ability to cope with the OOV problem and to directly preserve information from the source, which is important in this task.
Note that, as described in Section \ref{baseline}, we enhance the Pointer-Generator to become a no-template version of our model, therefore the effect of the head-modifier template can be measured by comparing the results of these two methods.
And the results demonstrate that our head-modifier template plays an important role in generating type descriptions.

In terms of the two proposed metrics, we find these metrics roughly positively correlated with traditional metrics, which in a way justifies our metrics.
These metrics provide interesting points of view on measuring generation quality.
The performance on \textbf{ModCopy} indicates that methods (Ptr-Gen, ours) with copy mechanism improves data fidelity by copying facts from the source, and the template helps the model know where and how to copy.
The performance on \textbf{HedAcc} demonstrates that our method is relatively better at predicting types for an entity, which in a way suggests the templates help the generated text maintain the head-modifier structure so that the head word is successfully parsed by the dependency parsing technique.
Although, we notice that in \textbf{Wiki200K}, models perform relatively worse on \textbf{ModCopy} and better on \textbf{HedAcc} than in \textbf{Wiki10K}.
This is most likely because the types of entities are finite, and more training data leads to more accuracy in predicting types.
Due to the size of the dataset and the limit of vocabulary size, the factual information is harder to preserve in the output.
This again proves the necessity of the new dataset.

\subsubsection{Manual Evaluation}

In this task, the readability of the generated type description is mostly related to its grammatical correctness, which benefits from the head-modifier templates.
Therefore, in order to measure the influence the templates make in terms of readability as well as how \textbf{ModCopy} (\textbf{M.C.}) and \textbf{HedAcc} (\textbf{H.A.}) correlate with manual judgment, we manually evaluate the generation from two aspects: \textit{Grammar Accuracy} (G.A.) and \textit{Overall Accuracy} (O.A.).
In detail, \textit{Grammar Accuracy} is the grammatical correctness judging by the grammar of the generated text alone; \textit{Overall Accuracy} is the grammatical and de facto correctness of the generated type description given an infobox and the ground truth. 
Note that \textit{Overall Accuracy} is always lower than or equal to \textit{Grammar Accuracy}.

In our experiment, we randomly select 200 pieces of data from the test set of \textbf{Wiki200K}, and provide the results of each method to the volunteers (who are all undergraduates) for manual evaluation.
We make sure each result is evaluated by two volunteers so as to eliminate the influence of subjective factors to some extent.

\begin{table}[h]
    \centering
	\begin{tabular}{ccccc}
		\toprule
        \textbf{Model}&\textit{G.A.}&\textit{O.A.}&\textbf{M.C.}&\textbf{H.A.}\\
		\midrule
		AttnS2S & 92.25 & 50.50 & 51.53 & 80.27 \\
		Ptr-Gen & 90.00 & 65.00 & \textbf{62.50} & 88.01 \\
		Transformer & 95.25 & 58.00 & 55.70 & 89.67 \\
		DGN & 89.50 & 56.00 & 47.29 & 81.37 \\
		\textbf{Our work} & \textbf{96.50} & \textbf{66.25} & 61.32 & \textbf{90.29} \\
        \bottomrule
	\end{tabular}
	\caption{Results of manual evaluation as well as two proposed metrics.}
	\label{table:manual}
\end{table}

The results, as shown in Table \ref{table:manual}, prove again the effectiveness of our method.
Our method outperforms other baselines in term of \textit{Grammar Accuracy}, which demonstrates that the model benefits from the head-modifier templates in term of readability by knowing ``how to say it''. 
In particular, the templates improves the \textit{Grammar Accuracy} substantially compared with Ptr-Gen. 
Results on the \textit{Overall Accuracy} indicate that our method ensures readability as well as data-fidelity, which indicates that the model benefits from the templates by knowing ``what to say''.
As for the proposed metrics \textbf{ModCopy} and \textbf{HedAcc}, 
they are, in line with intuition, relatively positively correlated with human judgment in general.
Also, notice that the statistics on both metrics are consistent with Table \ref{table:evaltable}.

\subsubsection{Effect of Templates}

% \begin{table}[!h]
% 	\centering
% 	\begin{tabular}{|c|c|}
% 		\hline
% 		EntityID&Q859415\\
% 		\hline
% 		Gold&\textbf{commune} in paris, france\\
% 		\hline
% 		Gen. tmpl.&\$hed\$ in \$mod\$, \$mod\$\\
% 		\hline
% 		Gen. desc.&\textbf{commune} in paris, france\\
% 		\hline
% 		Repl. tmpl. \#1 &\$mod\$ \$hed\$\\
% 		\hline
% 		Output \#1&\textbf{commune} in france\\
% 		\hline
% 		Repl. tmpl. \#2&\$hed\$ \$mod\$\\
% 		\hline
% 		Output \#2&\textbf{commune}\\
% 		\hline
% 		\hline
% 		EntityID&Q18758590\\
% 		\hline
% 		Gold&italian \textbf{architect} and \textbf{teacher}\\
% 		\hline
% 		Gen. tmpl.&\$mod\$ \$hed\$ and \$hed\$\\
% 		\hline
% 		Gen. desc.&italian \textbf{architect} and \textcolor{red}{\textbf{architect}}\\
% 		\hline
% 		Repl. tmpl. \#1&\$mod\$ \$hed\$\\
% 		\hline
% 		Output \#1&italian \textbf{architect}\\
% 		\hline
% 		Repl. tmpl. \#2&\$hed\$ \$mod\$ and \$mod\$\\
% 		\hline
% 		Output \#2&\textcolor{red}{\textbf{italy}} and \textbf{teacher}\\
% 		\hline
% 	\end{tabular}
% 	\caption{Examples of replacing templates, where ``Gen.'', ``Repl.'', ``tmpl.'' and ``desc.'' are short for ``generated'', ``replaced'', ``template'' and ``type description''. We use \textbf{bold} to denote the heads and use \textcolor{orange}{red} to denote mistaken words.}
% 	\label{table:tmpleffect}
% \end{table}

We aim to investigate whether the model is able to correct itself if the template generated in Stage 1 deviates from the correct one.
We select cases from \textbf{Wiki10K} test set to conduct experiments.
During inference, we deliberately replace the template in Stage 2 to see if the generated text still complies with the given template or if the model will be able to generate the right type description.

\begin{figure}[h]
	\centering
	\includegraphics[width=\linewidth]{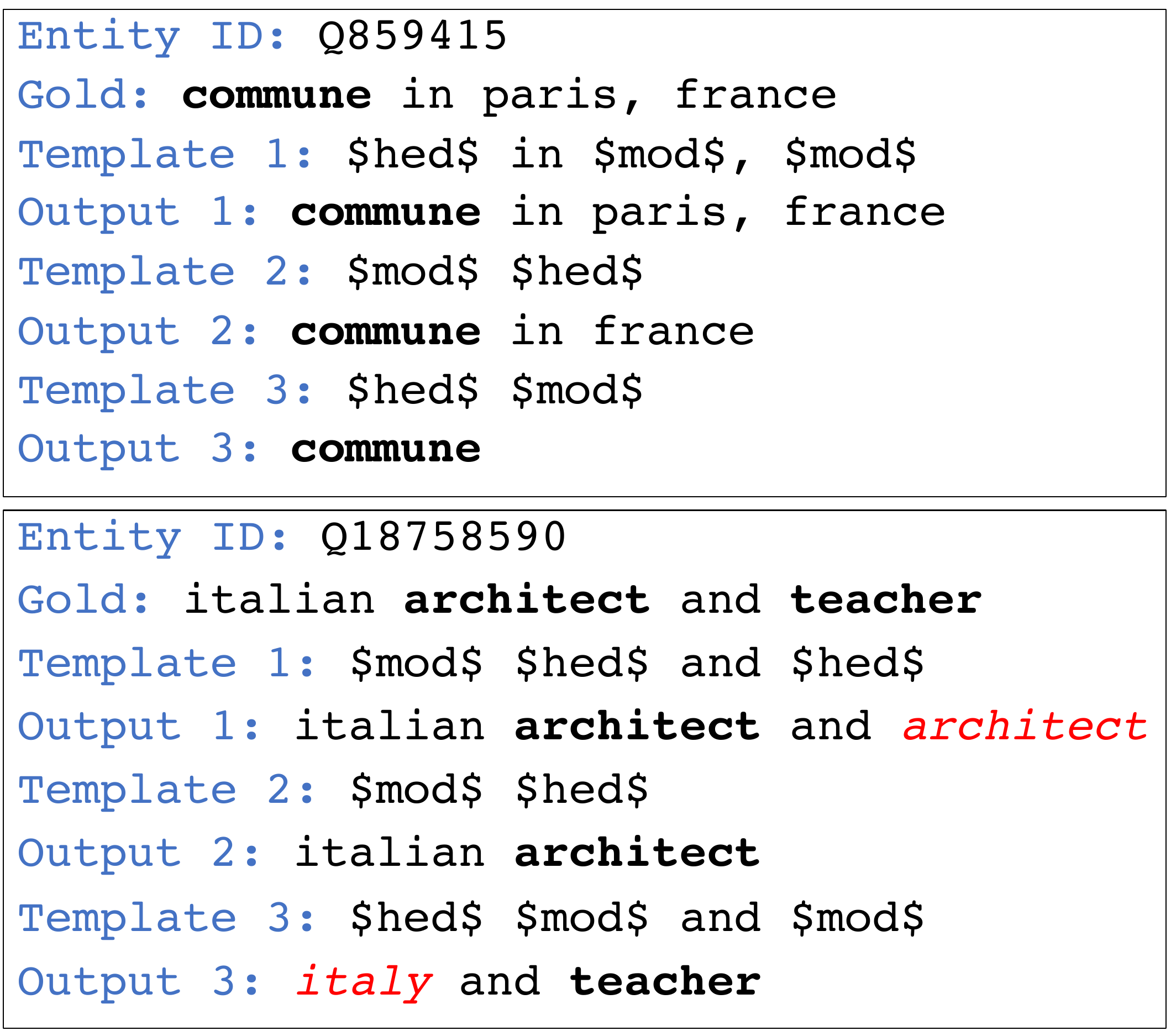}
	\caption{Examples of replacing templates. \texttt{Template 1}'s are the inital generated templates, while the remaining ones are produced by the authors. We use \textbf{bold} to denote the heads and use \textcolor{red}{\textit{italic red}} to denote mistaken words.}
	\label{fig:tmpleffect}
\end{figure}

The experimental results, as presented in Fig. \ref{fig:tmpleffect}, show our method's resilience against mistaken templates.
In the first case: 1) the replaced template \texttt{Template 2} is obviously inconsistent with the golden template \texttt{Template 1} (though it's also a possible template for other type descriptions), yet the model still manages to generate a type description though \texttt{paris} is lost; 
2) \texttt{Template 3} doesn't have the conjunction \texttt{in}, which causes confusion but the model still successfully predicts the right head.

In the second case, the model originally generates repetitive heads: 1) in \texttt{Template 2}, we delete the second \texttt{\$hed\$} in \texttt{Template 1}, and as a result, the model successfully generates a correct though incomplete output;
2) while \texttt{Template 3} is completely wrong judging by the head-modifier rule, and as a result \texttt{Output 3} is lost in readability. 
Nevertheless, due to the fact that the number of type descriptions is infinite yet the number of head-modifier templates is rather finite, the model can hardly generate a template that's completely wrong, therefore this scenario rarely happens in real life.
Still, the model tries to maintain a similar structure and successfully keeps data fidelity by predicting \texttt{teacher}, and preserving \texttt{italy}.

\section{Related Work}

There has been extensive work on mining entity-type pairs (i.e. isA relations) automatically. 
\citet{hearst1992automatic} uses a pattern-based method to extract isA pairs directly from free text with \textit{Hearst Patterns} (e.g., \textit{$NP_1$ is a $NP_2$}; \textit{$NP_0$ such as \{$NP_1, NP_2, ..., (and|or)\} NP_n$}) from which taxonomies can be induced \cite{poon2010unsupervised,velardi2013ontolearn,bansal2014structured}.
But these methods are limited in patterns, which often results in low recall and precision.

The most related line of work regarding predicting types for entities is entity-typing \cite{collins1999unsupervised,jiang2006exploiting,ratinov2009design}, which aims to assign types such as \texttt{people}, \texttt{location} from a fixed set to entity mentions in a document, and most of them model it a classification task. 
However, the types, even for those aiming at fine-grained entity-typing \cite{shimaoka2016neural,ren2016afet,anand2017fine} are too coarse-grained to be informative about the entity.
Also, the type set is too small and inflexible to meet the need for an ever-expanding KG.

In this task, the structured infobox is a source more suitable than textural data compared with text summarization task \cite{gu2016incorporating,See:2017wf,cao2018retrieve}, because not every entity in a KG possesses a paragraph of description.
For example, in CN-DBpedia \cite{xu2017cn}, which is one of the biggest Chinese KG, only a quarter of the entities have textual descriptions, yet almost every entity has an infobox.

Natural language generation (NLG) from structured data is a classic problem, in which many efforts have been made.
A common approach is to use hand-crafted templates \cite{kukich1983design,mckeown1992text}, but the acquisition of these templates in a specific domain is too costly.
Some also focus on automatically creating templates by clustering sentences and then use hand-crafted rules to induce templates \cite{angeli2010simple,konstas2013global}.
Recently with the rise of neural networks, many methods generate text in an end-to-end manner \cite{liu2017table,Wiseman:2017dp,Bhowmik2018EntityDescriptions}. 
However, they pay little attention to the grammatical structure of the output which may be ignored in generating long sentences, but it is crucial in generating short noun compounds like type descriptions.

\section{Conclusion and Future Work}

In this paper, we propose a head-modifier template-based type description generation method, powered by a copy mechanism and context gating mechanism.
We also propose a larger dataset and two metrics designed for this task.
Experimental results demonstrate that our method achieves state-of-the-art performance over baselines on both datasets while ensuring data fidelity and readability in generated type descriptions.
Further experiments regarding the effect of templates show that our model is not only controllable through templates, but resilient against wrong templates and able to correct itself.
Aside from such syntax templates, in the future, we aim to explore how semantic templates contribute to type description generation. 

\section{Acknowledgements}

We thank the anonymous reviewers for valuable comments.
This work was supported by National Key R\&D Program of China (No.2017YFC0803700), Shanghai Municipal Science and Technology Major Project (Grant No 16JC1420400).

\newpage
\bibliography{acl2019}

\begin{thebibliography}{40}
\expandafter\ifx\csname natexlab\endcsname\relax\def\natexlab#1{#1}\fi

\bibitem[{Anand et~al.(2017)Anand, Awekar et~al.}]{anand2017fine}
Ashish Anand, Amit Awekar, et~al. 2017.
\newblock Fine-grained entity type classification by jointly learning
  representations and label embeddings.
\newblock \emph{arXiv preprint arXiv:1702.06709}.

\bibitem[{Angeli et~al.(2010)Angeli, Liang, and Klein}]{angeli2010simple}
Gabor Angeli, Percy Liang, and Dan Klein. 2010.
\newblock A simple domain-independent probabilistic approach to generation.
\newblock In \emph{Proceedings of the 2010 Conference on Empirical Methods in
  Natural Language Processing}, pages 502--512. Association for Computational
  Linguistics.

\bibitem[{Auer et~al.(2007)Auer, Bizer, Kobilarov, Lehmann, Cyganiak, and
  Ives}]{auer2007dbpedia}
S{\"o}ren Auer, Christian Bizer, Georgi Kobilarov, Jens Lehmann, Richard
  Cyganiak, and Zachary Ives. 2007.
\newblock Dbpedia: A nucleus for a web of open data.
\newblock In \emph{The semantic web}, pages 722--735. Springer.

\bibitem[{Banerjee and Lavie(2005)}]{banerjee2005meteor}
Satanjeev Banerjee and Alon Lavie. 2005.
\newblock Meteor: An automatic metric for mt evaluation with improved
  correlation with human judgments.
\newblock In \emph{Proceedings of the acl workshop on intrinsic and extrinsic
  evaluation measures for machine translation and/or summarization}, pages
  65--72.

\bibitem[{Bansal et~al.(2014)Bansal, Burkett, De~Melo, and
  Klein}]{bansal2014structured}
Mohit Bansal, David Burkett, Gerard De~Melo, and Dan Klein. 2014.
\newblock Structured learning for taxonomy induction with belief propagation.
\newblock In \emph{Proceedings of the 52nd Annual Meeting of the Association
  for Computational Linguistics (Volume 1: Long Papers)}, volume~1, pages
  1041--1051.

\bibitem[{Bhowmik and {de Melo}(2018)}]{Bhowmik2018EntityDescriptions}
Rajarshi Bhowmik and Gerard {de Melo}. 2018.
\newblock Generating fine-grained open vocabulary entity type descriptions.
\newblock In \emph{Proceedings of ACL 2018}.

\bibitem[{Cao et~al.(2018)Cao, Li, Li, and Wei}]{cao2018retrieve}
Ziqiang Cao, Wenjie Li, Sujian Li, and Furu Wei. 2018.
\newblock Retrieve, rerank and rewrite: Soft template based neural
  summarization.
\newblock In \emph{Proceedings of the 56th Annual Meeting of the Association
  for Computational Linguistics (Volume 1: Long Papers)}, volume~1, pages
  152--161.

\bibitem[{Chung et~al.(2014)Chung, Gulcehre, Cho, and
  Bengio}]{chung2014empirical}
Junyoung Chung, Caglar Gulcehre, KyungHyun Cho, and Yoshua Bengio. 2014.
\newblock Empirical evaluation of gated recurrent neural networks on sequence
  modeling.
\newblock \emph{arXiv preprint arXiv:1412.3555}.

\bibitem[{Collins and Singer(1999)}]{collins1999unsupervised}
Michael Collins and Yoram Singer. 1999.
\newblock Unsupervised models for named entity classification.
\newblock In \emph{1999 Joint SIGDAT Conference on Empirical Methods in Natural
  Language Processing and Very Large Corpora}.

\bibitem[{Gu et~al.(2016)Gu, Lu, Li, and Li}]{gu2016incorporating}
Jiatao Gu, Zhengdong Lu, Hang Li, and Victor~OK Li. 2016.
\newblock Incorporating copying mechanism in sequence-to-sequence learning.
\newblock \emph{arXiv preprint arXiv:1603.06393}.

\bibitem[{Gulcehre et~al.(2016)Gulcehre, Ahn, Nallapati, Zhou, and
  Bengio}]{gulcehre2016pointing}
Caglar Gulcehre, Sungjin Ahn, Ramesh Nallapati, Bowen Zhou, and Yoshua Bengio.
  2016.
\newblock Pointing the unknown words.
\newblock \emph{arXiv preprint arXiv:1603.08148}.

\bibitem[{Hearst(1992)}]{hearst1992automatic}
Marti~A Hearst. 1992.
\newblock Automatic acquisition of hyponyms from large text corpora.
\newblock In \emph{Proceedings of the 14th conference on Computational
  linguistics-Volume 2}, pages 539--545. Association for Computational
  Linguistics.

\bibitem[{Hippisley et~al.(2005)Hippisley, Cheng, and
  Ahmad}]{hippisley2005head}
Andrew Hippisley, David Cheng, and Khurshid Ahmad. 2005.
\newblock The head-modifier principle and multilingual term extraction.
\newblock \emph{Natural Language Engineering}, 11(2):129--157.

\bibitem[{Jiang and Zhai(2006)}]{jiang2006exploiting}
Jing Jiang and ChengXiang Zhai. 2006.
\newblock Exploiting domain structure for named entity recognition.
\newblock In \emph{Proceedings of the main conference on Human Language
  Technology Conference of the North American Chapter of the Association of
  Computational Linguistics}, pages 74--81. Association for Computational
  Linguistics.

\bibitem[{Kingma and Ba(2014)}]{kingma2014adam}
Diederik~P Kingma and Jimmy Ba. 2014.
\newblock Adam: A method for stochastic optimization.
\newblock \emph{arXiv preprint arXiv:1412.6980}.

\bibitem[{Klein et~al.(2017)Klein, Kim, Deng, Senellart, and Rush}]{opennmt}
Guillaume Klein, Yoon Kim, Yuntian Deng, Jean Senellart, and Alexander~M. Rush.
  2017.
\newblock \href {https://doi.org/10.18653/v1/P17-4012} {Open{NMT}: Open-source
  toolkit for neural machine translation}.
\newblock In \emph{Proc. ACL}.

\bibitem[{Konstas and Lapata(2013)}]{konstas2013global}
Ioannis Konstas and Mirella Lapata. 2013.
\newblock A global model for concept-to-text generation.
\newblock \emph{Journal of Artificial Intelligence Research}, 48:305--346.

\bibitem[{Kukich(1983)}]{kukich1983design}
Karen Kukich. 1983.
\newblock Design of a knowledge-based report generator.
\newblock In \emph{Proceedings of the 21st annual meeting on Association for
  Computational Linguistics}, pages 145--150. Association for Computational
  Linguistics.

\bibitem[{Lebret et~al.(2016)Lebret, Grangier, and Auli}]{lebret2016neural}
R{\'e}mi Lebret, David Grangier, and Michael Auli. 2016.
\newblock Neural text generation from structured data with application to the
  biography domain.
\newblock \emph{arXiv preprint arXiv:1603.07771}.

\bibitem[{Lin(2004)}]{lin2004rouge}
Chin-Yew Lin. 2004.
\newblock Rouge: A package for automatic evaluation of summaries.
\newblock \emph{Text Summarization Branches Out}.

\bibitem[{Liu et~al.(2017)Liu, Wang, Sha, Chang, and Sui}]{liu2017table}
Tianyu Liu, Kexiang Wang, Lei Sha, Baobao Chang, and Zhifang Sui. 2017.
\newblock Table-to-text generation by structure-aware seq2seq learning.
\newblock \emph{arXiv preprint arXiv:1711.09724}.

\bibitem[{Luong et~al.(2015)Luong, Pham, and Manning}]{luong2015effective}
Minh-Thang Luong, Hieu Pham, and Christopher~D Manning. 2015.
\newblock Effective approaches to attention-based neural machine translation.
\newblock \emph{arXiv preprint arXiv:1508.04025}.

\bibitem[{Manning et~al.(2014)Manning, Surdeanu, Bauer, Finkel, Bethard, and
  McClosky}]{manning2014stanford}
Christopher Manning, Mihai Surdeanu, John Bauer, Jenny Finkel, Steven Bethard,
  and David McClosky. 2014.
\newblock The stanford corenlp natural language processing toolkit.
\newblock In \emph{Proceedings of 52nd annual meeting of the association for
  computational linguistics: system demonstrations}, pages 55--60.

\bibitem[{McKeown(1992)}]{mckeown1992text}
Kathleen McKeown. 1992.
\newblock \emph{Text generation}.
\newblock Cambridge University Press.

\bibitem[{Papineni et~al.(2002)Papineni, Roukos, Ward, and
  Zhu}]{papineni2002bleu}
Kishore Papineni, Salim Roukos, Todd Ward, and Wei-Jing Zhu. 2002.
\newblock Bleu: a method for automatic evaluation of machine translation.
\newblock In \emph{Proceedings of the 40th annual meeting on association for
  computational linguistics}, pages 311--318. Association for Computational
  Linguistics.

\bibitem[{Poon and Domingos(2010)}]{poon2010unsupervised}
Hoifung Poon and Pedro Domingos. 2010.
\newblock Unsupervised ontology induction from text.
\newblock In \emph{Proceedings of the 48th annual meeting of the Association
  for Computational Linguistics}, pages 296--305. Association for Computational
  Linguistics.

\bibitem[{Ratinov and Roth(2009)}]{ratinov2009design}
Lev Ratinov and Dan Roth. 2009.
\newblock Design challenges and misconceptions in named entity recognition.
\newblock In \emph{Proceedings of the Thirteenth Conference on Computational
  Natural Language Learning}, pages 147--155. Association for Computational
  Linguistics.

\bibitem[{Ren et~al.(2016)Ren, He, Qu, Huang, Ji, and Han}]{ren2016afet}
Xiang Ren, Wenqi He, Meng Qu, Lifu Huang, Heng Ji, and Jiawei Han. 2016.
\newblock Afet: Automatic fine-grained entity typing by hierarchical
  partial-label embedding.
\newblock In \emph{Proceedings of the 2016 Conference on Empirical Methods in
  Natural Language Processing}, pages 1369--1378.

\bibitem[{Schuster and Paliwal(1997)}]{schuster1997bidirectional}
Mike Schuster and Kuldip~K Paliwal. 1997.
\newblock Bidirectional recurrent neural networks.
\newblock \emph{IEEE Transactions on Signal Processing}, 45(11):2673--2681.

\bibitem[{See et~al.(2017)See, Liu, and Manning}]{See:2017wf}
Abigail See, Peter~J Liu, and Cristopher~D Manning. 2017.
\newblock {Get To The Point: Summarization with Pointer-Generator Networks }.
\newblock pages 1--20.

\bibitem[{Sha et~al.(2018)Sha, Mou, Liu, Poupart, Li, Chang, and
  Sui}]{sha2018order}
Lei Sha, Lili Mou, Tianyu Liu, Pascal Poupart, Sujian Li, Baobao Chang, and
  Zhifang Sui. 2018.
\newblock Order-planning neural text generation from structured data.
\newblock In \emph{Thirty-Second AAAI Conference on Artificial Intelligence}.

\bibitem[{Shimaoka et~al.(2016)Shimaoka, Stenetorp, Inui, and
  Riedel}]{shimaoka2016neural}
Sonse Shimaoka, Pontus Stenetorp, Kentaro Inui, and Sebastian Riedel. 2016.
\newblock Neural architectures for fine-grained entity type classification.
\newblock \emph{arXiv preprint arXiv:1606.01341}.

\bibitem[{Tu et~al.(2017)Tu, Liu, Lu, Liu, and Li}]{tu2017context}
Zhaopeng Tu, Yang Liu, Zhengdong Lu, Xiaohua Liu, and Hang Li. 2017.
\newblock Context gates for neural machine translation.
\newblock \emph{Transactions of the Association for Computational Linguistics},
  5:87--99.

\bibitem[{Vaswani et~al.(2017)Vaswani, Shazeer, Parmar, Uszkoreit, Jones,
  Gomez, Kaiser, and Polosukhin}]{vaswani2017attention}
Ashish Vaswani, Noam Shazeer, Niki Parmar, Jakob Uszkoreit, Llion Jones,
  Aidan~N Gomez, {\L}ukasz Kaiser, and Illia Polosukhin. 2017.
\newblock Attention is all you need.
\newblock In \emph{Advances in Neural Information Processing Systems}, pages
  5998--6008.

\bibitem[{Vedantam et~al.(2015)Vedantam, Lawrence~Zitnick, and
  Parikh}]{vedantam2015cider}
Ramakrishna Vedantam, C~Lawrence~Zitnick, and Devi Parikh. 2015.
\newblock Cider: Consensus-based image description evaluation.
\newblock In \emph{Proceedings of the IEEE conference on computer vision and
  pattern recognition}, pages 4566--4575.

\bibitem[{Velardi et~al.(2013)Velardi, Faralli, and
  Navigli}]{velardi2013ontolearn}
Paola Velardi, Stefano Faralli, and Roberto Navigli. 2013.
\newblock Ontolearn reloaded: A graph-based algorithm for taxonomy induction.
\newblock \emph{Computational Linguistics}, 39(3):665--707.

\bibitem[{Vrande{\v{c}}i{\'c} and Kr{\"o}tzsch(2014)}]{vrandevcic2014wikidata}
Denny Vrande{\v{c}}i{\'c} and Markus Kr{\"o}tzsch. 2014.
\newblock Wikidata: a free collaborative knowledgebase.
\newblock \emph{Communications of the ACM}, 57(10):78--85.

\bibitem[{Wang et~al.(2014)Wang, Wang, and Hu}]{wang2014head}
Zhongyuan Wang, Haixun Wang, and Zhirui Hu. 2014.
\newblock Head, modifier, and constraint detection in short texts.
\newblock In \emph{2014 IEEE 30th International Conference on Data
  Engineering}, pages 280--291. IEEE.

\bibitem[{Wiseman et~al.(2017)Wiseman, Shieber, and Rush}]{Wiseman:2017dp}
Sam Wiseman, Stuart Shieber, and Alexander Rush. 2017.
\newblock {Challenges in Data-to-Document Generation}.
\newblock In \emph{Proceedings of the 2017 Conference on Empirical Methods in
  Natural Language Processing}, pages 2253--2263, Stroudsburg, PA, USA.
  Association for Computational Linguistics.

\bibitem[{Xu et~al.(2017)Xu, Xu, Liang, Xie, Liang, Cui, and Xiao}]{xu2017cn}
Bo~Xu, Yong Xu, Jiaqing Liang, Chenhao Xie, Bin Liang, Wanyun Cui, and Yanghua
  Xiao. 2017.
\newblock Cn-dbpedia: A never-ending chinese knowledge extraction system.
\newblock In \emph{International Conference on Industrial, Engineering and
  Other Applications of Applied Intelligent Systems}, pages 428--438. Springer.

\end{thebibliography}
\bibliographystyle{acl_natbib}

\end{document}